\documentclass{article}
\usepackage{amsmath,mlspconf}
\usepackage{amssymb}

\usepackage{graphicx}
\graphicspath{ {fig/} } 

\usepackage{float}

\usepackage{algorithm,algorithmic}

\usepackage{url}
\usepackage[]{hyperref}
\usepackage{breakurl}

\usepackage{color}

\copyrightnotice{978-1-5090-0746-2/16/\$31.00 {\copyright}2016 IEEE}

\toappear{2016 IEEE International Workshop on Machine Learning for Signal Processing, Sept.\ 13--16, 2016, Salerno, Italy}

\newcommand{\dataset}{{\cal D}}
\newcommand{\GP}[0]{\mathcal{GP}}
\newcommand{\Normal}[0]{\mathcal{N}}

\newcommand{\x}{\mathbf{x}}

\newcommand{\f}{\mathbf{f}}

\newcommand{\h}{\mathbf{h}}
\newcommand{\q}{\mathbf{q}}
\newcommand{\Q}{\mathbf{Q}}

\newcommand{\kt}{\vect{k}_{t+1}}

\newcommand{\K}{\mathbf{K}}

\newcommand{\I}{\mathbf{I}}

\newcommand{\zeros}{\mathbf{0}}

\newcommand{\Real}[0]{\mathbb{R}}

\newcommand{\vect}[1]{{\boldsymbol{\mathbf{#1}}}} 
\newcommand{\mat}[1]{{\boldsymbol{\mathbf{#1}}}} 
\newcommand{\pmean}{\vect{\mu}}
\newcommand{\pcov}{\mat{\Sigma}}

\title{On the Relationship between Online Gaussian Process Regression\\ and Kernel Least Mean Squares Algorithms}
\name{Steven Van Vaerenbergh$^\star$, Jesus Fernandez-Bes$^\dagger$ $^\ddagger$, V\'ictor Elvira$^\ast$
    \thanks{The work of S. Van Vaerenbergh is supported by the Spanish Ministry of Economy and Competitiveness, under projects PRISMA (TEC2014-57402-JIN) and RACHEL (TEC2013-47141-C4-3-R).\newline \-\hspace{1.5em} J. Fernandez-Bes' work was partially supported by projects TIN2013-41998-R, TEC2014-52289-R, and PRICAM S2013/ICE-2933. \newline \-\hspace{1.5em} The work of V. Elvira is supported by the Spanish Ministry of Economy and Competitiveness, under project TEC2013-41718-R.}
}
\address{
    $^\star$ Dept. of Communications Engineering, University of Cantabria, Spain\\
    $^\dagger$ CIBER-BBN, Zaragoza, Spain,  \\ 
    $^\ddagger$ BSICoS Group, I3A, IIS Arag\'on, University of Zaragoza, Zaragoza, Spain.\\
     $^\ast$ Dept. of Signal Theory and Communications, Universidad Carlos III de Madrid, Spain\\
}

\begin{document}
%

\maketitle
\begin{abstract}
We study the relationship between online Gaussian process (GP) regression and kernel least mean squares (KLMS) algorithms. While the latter have no capacity of storing the entire posterior distribution during online learning, we discover that their operation corresponds to the assumption of a fixed posterior covariance that follows a simple parametric model. Interestingly, several well-known KLMS algorithms correspond to specific cases of this model. The probabilistic perspective allows us to understand how each of them handles uncertainty, which could explain some of their performance differences. 




\end{abstract}


\section{Introduction}
\label{sec:intro}
Gaussian Process (GP) regression is a state-of-the-art Bayesian technique for nonlinear regression \cite{rasmussen2006gaussian}. Although GP models were proposed in the seventies \cite{o1978curve}, they did not become widely applied tools in machine learning until the last decade, mainly due to their computational complexity. 

Through what is known as the ``kernel trick'', GP regression extends least squares to nonlinear estimation. By doing so, GP regression can be considered the natural Bayesian nonlinear extension of linear minimum mean square error estimation (MMSE) algorithms, which are central in signal processing \cite{perezcruz2013gaussian}. Closely related to GPs are kernel methods \cite{scholkopf2002learning}, which have been successfully applied to several nonlinear signal processing problems, such as classification with support vector machines and kernel PCA for nonlinear dimensionality reduction. The main difference between Bayesian methods such as GPs and kernel methods is that the former provide a full probability distribution of the estimated variables, while the latter obtain only a point estimate.

Several kernel extensions of classical adaptive filters have been proposed in the literature (see for instance \cite{liu2010kernel,vanvaerenbergh2014online} and the references therein). These algorithms, referred to as \emph{kernel adaptive filtering} are mainly divided into two families, similar to the linear adaptive filtering literature: (i) kernel least-mean-squares (KLMS) algorithms \cite{liu2008kernel,richard2009online,chen2012quantized}, which are based on stochastic gradient minimization of the mean square error and have linear complexity per iteration w.r.t. the number of data points; (ii) and kernel recursive-least-squares (KRLS) algorithms \cite{engel2004kernel}, which recursively solve the least-squares problem, using quadratic complexity per iteration.

In \cite{vanvaerenbergh2012kernel}, an online formulation of GP regression was obtained by deriving KRLS from a Bayesian point of view. An equivalent formulation for online GPs was presented in \cite{csato02sparse}, though we will follow \cite{vanvaerenbergh2012kernel} as it offers a more intuitive choice for the variables and it provides a direct connection with KRLS algorithms. The online GP formulation from \cite{vanvaerenbergh2012kernel} adds two notable features to the KRLS literature: it allows the use of maximization techniques to set the hyperparameters without using cross-validation, and it provides an uncertainty measurement of the estimate.

KLMS algorithms are much more popular than KRLS, due to their low complexity. Interestingly though, as far as we know, there has not been a similar fully probabilistic interpretation of KLMS algorithms. Note that there exist some Bayesian interpretations of the LMS algorithm \cite{fernandez2015probabilistic,huemmer2015nlms}, one of which considers kernels, albeit in a simplified setting \cite{park2014probabilistic}.

In this work we provide a novel derivation of KLMS starting from a Bayesian model based on GPs. Using the sequential update rule of online GPs and a systematic approximation of its posterior covariance matrix, we are able to derive a KLMS formulation that generalizes the two main KLMS formulations, namely the KLMS algorithm \cite{liu2008kernel} and the KNLMS algorithm \cite{richard2009online}. The connection we establish with Gaussian processes sheds new light on the manner in which KLMS algorithms deal with uncertainty.





\section{Online GP regression}
\label{sec:ogp}
\subsection{Gaussian process regression}

Consider a set of $N$ input-output pairs $\dataset = \{\vect{x}_i, y_i\}_{i=1}^N$, where $\vect{x}_i \in \Real^D$ are $D$-dimensional input vectors and $y_i\in\Real$ are scalar outputs. We assume that the observed data can be described by the following model,
\begin{equation}
y_i = f(\x_i) + \varepsilon_i,
\label{eq:regmodel}
\end{equation}
in which $f$ represents an unobservable \emph{latent function} and $\varepsilon_i \sim \mathcal{N}(0,\sigma_n^2)$ is zero-mean Gaussian noise.

A Gaussian process is a collection of random variables, any finite number of which have a joint Gaussian distribution \cite{rasmussen2006gaussian}. 
To indicate that a random function $f(\x)$ follows a Gaussian process we write it as
\begin{equation*}
f(\x) \sim \GP(m(\x),k(\x,\x')).
\end{equation*}
All values of $f$ at any locations $\x$ are jointly normally distributed, with $m(\x)$ and $k(\x,\x')$ representing the mean function and covariance function, respectively.

In a Bayesian regression setting, we are interested in inferring the predictive distribution of a new, unseen output $y_{\ast}$ given the corresponding input $\x_{\ast}$ and the data $\dataset$.
%
%
In particular, we take a Gaussian process as the prior over the latent function, and the vector of observations $[y_1,\dots,y_n]^\top$ is related to the latent function through the likelihood function $p(\vect{y}|f)$. When the observations are contaminated with zero-mean Gaussian noise, as in Eq.~\eqref{eq:regmodel},
there exists a closed-form solution for the \emph{posterior distribution} over functions, i.e. the distribution over the unknown function $f(x)$ after incorporating all the observed data. Specifically, the posterior  of the function at any new location $\x_{\ast}$ is described by
\begin{equation*}
p(f_{\ast}|\x_{\ast},\dataset) = \Normal(\hat f_\ast,\hat \sigma_\ast^2).
\end{equation*}
When $m(\x)=0$, which is a very common assumption, we obtain the following expressions \cite{rasmussen2006gaussian}
\begin{subequations}
\begin{align}
\hat f_\ast &= \vect{k}_\ast^\top (\K + \sigma_n^2 \I)^{-1} \vect{y} \label{eq:pred_mean_gpml}\\
\hat \sigma_\ast^2 & = k_{\ast\ast} - \vect{k}_\ast^\top (\K + \sigma_n^2 \I)^{-1} \vect{k}_\ast^\top \label{eq:pred_var_gpml}
\end{align}
\label{eq:pred_distr_gpml}%
\end{subequations}
where the covariances (or \emph{kernel}) matrices $\K$ contain the elements $[\K]_{ij} = k(\x_i,\x_j)$ and we have introduced the shorthand notations $\vect{k}_\ast = [k(\x_1,\x_\ast),\dots,k(\x_N,\x_\ast)]^\top$ and $k_{\ast\ast} = k(\x_\ast,\x_\ast)$.
The matrix inversion involved in Eqs.~\eqref{eq:pred_mean_gpml} and \eqref{eq:pred_var_gpml} leads to $\mathcal{O}(N^3)$ complexity.

\subsection{Incremental GP updates}

In an \emph{online} scenario, the data pairs are made available on a one-at-a-time basis, i.e. $(\x_t,y_t)$ arrives at time $t$. Instead of recalculating the predictive distribution entirely once a new data pair $(\x_{t+1},y_{t+1})$ arrives, i.e. by solving \eqref{eq:pred_distr_gpml}, it is more interesting to perform an incremental update. 
In this section we briefly review the sequential updates for online GP regression as presented in \cite{vanvaerenbergh2012kernel}. In order to avoid the unbounded growth of the involved matrices, the online learning process is typically coupled with a \emph{sparsification} procedure.

At the $t$-th iteration of the online GP, the model contains the variables
\begin{equation*}
\mathcal{M}_t = \{\dataset_t,\pmean_t,\pcov_t,\Q_t\},
\end{equation*}
where $\dataset_t$ is the observed data set that contains the data pairs $\{\x_i,y_i\}_{i=1}^t$; $\pmean_t$ and $\pcov_t$ are the mean and covariance matrices of the posterior $p(\f_t|\dataset_t) = \Normal(\pmean_t,\pcov_t)$; and $\Q_t = \K_t^{-1}$ is the inverse covariance matrix corresponding to $\dataset_t$.

When a new data pair $(\x_{t+1},y_{t+1})$ is obtained, the posterior distribution
\begin{equation*}
p(\f_{t+1}|\dataset_{t+1}) = \Normal(\f_{t+1}| \pmean_{t+1}, \pcov_{t+1})
\end{equation*}
is updated as
\begin{subequations}
\begin{align}
\pmean_{t+1} & =
\begin{bmatrix} \pmean_t \\ \hat{y}_{t+1} \end{bmatrix}
+ \frac{y_{t+1} - \hat{y}_{t+1}}{\hat{\sigma}_{yt+1}^2}
\begin{bmatrix} \h_{t+1} \\ \hat{\sigma}_{ft+1}^2, \end{bmatrix}, \label{eq:pmean} \\
\pcov_{t+1} & = 
\begin{bmatrix} \pcov_t & \h_{t+1}\\ \h_{t+1}^\top & \hat{\sigma}_{ft+1}^2 \end{bmatrix}
- \frac{1}{\hat{\sigma}_{yt+1}^2}
\begin{bmatrix} \h_{t+1} \\ \hat{\sigma}_{ft+1}^2 \end{bmatrix}
\begin{bmatrix} \h_{t+1} \\ \hat{\sigma}_{ft+1}^2 \end{bmatrix}^\top, \label{eq:pcov}
\end{align}
\end{subequations}
where
\begin{align*}
\q_{t+1} & = \Q_t \kt, \\
\h_{t+1} & = \pcov_t \q_{t+1},
\end{align*}
and the vector $\kt$ has elements $[\kt]_i = k(\x_i,\x_{t+1})$. Furthermore, the output variance is obtained as
\begin{equation*}
\hat{\sigma}_{yt+1}^2 = \sigma_n^2 + \hat{\sigma}_{ft+1}^2,
\label{eq:sigma_y}
\end{equation*}
and the variance of the latent function evaluations is
\begin{align*}
\hat{\sigma}_{ft+1}^2 & = k_{t+1} + \kt^\top (\Q_t\pcov_t\Q_t-\Q_t)\kt\\
& = \gamma_{t+1}^2 + \q_{t+1}^T \h_{t+1}.
\end{align*}
In order to further simplify equations, the variable 
\begin{equation*}
\gamma_{t+1}^2 = k_{t+1} - \kt^T \Q_t \kt
\end{equation*}
is introduced.
The inverse kernel matrix $\Q_t$ can be updated efficiently through
\begin{equation}
\Q_{t+1} = 
\begin{bmatrix}
\Q_t & \zeros \\ \zeros^\top & 0
\end{bmatrix}
+ \frac{1}{\gamma_{t+1}^2}
\begin{bmatrix}
\q_{t+1} \\ -1
\end{bmatrix}
\begin{bmatrix}
\q_{t+1} \\ -1
\end{bmatrix}^\top.
\label{eq:q_update}
\end{equation}
After applying Eqs.~\eqref{eq:pmean}, \eqref{eq:pcov}, and \eqref{eq:q_update}, we obtain the updated model $\mathcal{M}_{t+1} = \{\dataset_{t+1},\pmean_{t+1},\pcov_{t+1},\Q_{t+1}\}$.

At each step $t$ of the learning process, the predictive distribution of a new observation $y_{t+1}$ given all past data is a Gaussian
%
$
p(y_{t+1}|\dataset_t) = \Normal(y_{t+1}|\hat y_{t+1},\hat \sigma_{yt+1}^2)
$
with
\begin{subequations}
\begin{align}
\hat y_{t+1} & = \q_{t+1}^\top \pmean_t = \kt^\top \Q_t \pmean_t \label{eq:pred_mean_ogp} \\
\hat \sigma_{yt+1}^2 & = \sigma_n^2 + k_{t+1} + 
\kt^\top (\Q_t\pcov_t\Q_t-\Q_t)\kt. \label{eq:pred_cov_ogp}
\end{align}
\label{eq:pred_distr_ogp}
\end{subequations}
%
%
For more details we refer the reader to \cite{vanvaerenbergh2012kernel,lazaro2011bayesian}.


\section{Kernel adaptive filtering and KLMS}
\label{sec:klms}
Kernel methods are a class of machine learning algorithms that are closely related to Gaussian processes. In many cases, kernel methods obtain the same solution as their GP counterpart. For instance, the most popular regression algorithm in kernel methods, kernel ridge regression (KRR) \cite{saunders2002string}, obtains Eq.~\eqref{eq:pred_mean_gpml} for predicting new outputs, which, in the kernel methods literature, is expressed as
\begin{equation}
\hat f_\ast = \vect{\alpha}^\top \vect{k}_\ast = \sum_{i=1}^N \alpha_i k(\x_i,\x_{\star}).
\end{equation}
Vector $\vect{\alpha}$ contains the ``kernel weights'', which are found as $\vect{\alpha}^\top = (\K + \sigma_n^2 \I)^{-1} \vect{y}$. Nonetheless, kernel methods do not follow a probabilistic Bayesian approach: their solution corresponds only to a point estimate, and they do not model the entire predictive distribution. This implies, among others, that kernel methods do not handle prediction uncertainty out-of-the-box. GP regression, in contrast, provides the predictive variance \eqref{eq:pred_var_gpml} in addition to the predictive mean.

\subsection{Kernel recursive least-squares}
The kernel-methods counterpart of online GP regression is kernel recursive least-squares (KRLS, see for instance \cite{engel2004kernel}), which obtains the KRR solution recursively. After receiving $t$ data points, the kernel weights obtained by KRLS are those that solve the batch problem
\begin{equation}
\vect{\alpha}_t = (\K_t + \sigma_n^2\mat{I})^{-1}\vect{y}_t.
\label{eq:krls_alpha}
\end{equation}
The same weights can be obtained in online GP regression by computing
\begin{equation}
\vect{\alpha}_t = \K_t^{-1}\pmean_t = \Q_t\pmean_t,
\end{equation}
which follows from Eq.~\eqref{eq:pred_mean_ogp}.

\subsection{KLMS algorithms} 

Similar to online GP regression, updating the KRLS estimate in Eq.~\eqref{eq:krls_alpha} with a new data point requires quadratic complexity. Kernel least-mean-squares (KLMS) algorithms alleviate this computational burden by performing stochastic gradient descent of the mean square error, resulting in linear complexity per time step \cite{vanvaerenbergh2014online}.

KLMS algorithms can be categorized into two classes, depending on which kernel weights they update in each iteration to account for the prediction error\footnote{Both approaches are different approximations of the same update formulation in the \emph{kernel feature space}; see \cite{vanvaerenbergh2014online} for details.}. We outline both approaches briefly in the remainder of this section.

\subsubsection{Type-I KLMS: concentrating the novelty}

The first type of KLMS algorithm updates only one coefficient in order to compensate for the prediction error, at each time step. When the weight vector is allowed to grow, this update takes the form
\begin{equation}
\vect{\alpha}_{t+1}^{(I)} = \begin{bmatrix}
\vect{\alpha}_t\\ \eta e_{t+1}
\end{bmatrix},
\label{eq:klms_update}
\end{equation}
in which $e_{t+1} = y_{t+1} - \hat y_{t+1}$ is the instantaneous error.

Eq.~\eqref{eq:klms_update} represents the basic update of what is known as the KLMS algorithm, proposed in \cite{liu2008kernel}. In order to avoid the infinite growth of $\vect{\alpha}_t$, a more sophisticated version of this algorithm was presented in \cite{chen2012quantized}, known as Quantized Kernel Least Mean Square (QKLMS). When QKLMS receives a datum similar to a previously seen datum, for instance the $i$-th base it has stored, it does not expand $\vect{\alpha}_t$ but instead updates the corresponding weight $\alpha_i$.

\subsubsection{Type-II KLMS: spreading the novelty}

A different strategy consists in updating all coefficients of $\vect{\alpha}_t$ in each iteration. This approach is followed for instance by the Kernel Normalized Least Mean Square (KNLMS) algorithm \cite{richard2009online}, whose update reads
\begin{equation}
\vect{\alpha}_{t+1}^{(II)} = \begin{bmatrix}
\vect{\alpha}_t\\ 0
\end{bmatrix} + \eta \frac{e_{t+1}}{\epsilon + k_{t+1}^2 + \|\kt\|^2} \begin{bmatrix}
\kt\\ k_{t+1}
\end{bmatrix}
\label{eq:knlms_update}
\end{equation}
when $\vect{\alpha}_t$ is allowed to grow. Note that the rule \eqref{eq:knlms_update} updates all coefficients in each iteration. In order to avoid unbounded growth, KNLMS follows a \emph{coherence} criterion that promotes sparsity. 


\section{From online GP to KLMS}
\label{sec:approx}
The update of the KLMS kernel weights, $\vect{\alpha}_{t}$, can be obtained in terms of the online GP's predictive mean, $\pmean_{t}$, by elaborating $\vect{\alpha}_{t+1} = \Q_{t+1}\pmean_{t+1}$, which results in
\begin{equation}
\vect{\alpha}_{t+1} = \begin{bmatrix}
\vect{\alpha}_t\\ 0
\end{bmatrix}
+ \frac{e_{t+1}}{\hat{\sigma}_{yt+1}^2}
\begin{bmatrix}
(\Q_t \pcov_t \Q_t - \Q_t) \kt\\ 1
\end{bmatrix}.
\label{eq:update_alpha}
\end{equation}
Comparison of Eq.~\eqref{eq:update_alpha} with Eqs.~\eqref{eq:klms_update} and \eqref{eq:knlms_update} indicates that, in order to obtain a KLMS-like update rule, the covariance $\pcov_t$ is to be replaced by the following parametric model:
\begin{equation}
\pcov_t = \K_t\left(\beta\K_t + \I\right),
\label{eq:pcov_simple}
\end{equation}
which implies
\begin{equation}
\Q_t \pcov_t \Q_t - \Q_t = {\beta}\I.
\end{equation}
This substitution indicates that, instead of Eq.~\eqref{eq:pred_var_gpml}, the following predictive variance is assumed
\begin{equation}
\hat\sigma^2_{ft+1} = k_{t+1} + {\beta}\|\kt\|^2.\label{eq:beta_sigma}
\end{equation}
The update of the predictive mean, Eq.~\eqref{eq:update_alpha}, then simplifies to an expression that does not contain $\pcov_t$ nor $\Q_t$,
\begin{equation}
\vect{\alpha}_{t+1}^{\beta} = \begin{bmatrix}
\vect{\alpha}_t\\ 0
\end{bmatrix} + \frac{e_{t+1}}{\sigma_n^2 + k_{t+1} + \beta\|\kt\|^2} \begin{bmatrix}
\beta\kt\\ 1
\end{bmatrix}.
\label{eq:update_alpha_subs}
\end{equation}
The update rule of Eq.~\eqref{eq:update_alpha_subs} has linear complexity with respect to the number of processed data points, $t+1$. 

\subsection{Specific cases: $\beta=0$ and $\beta=1$}

Setting $\beta=0$ in Eq.~\eqref{eq:update_alpha_subs} yields the update rule
\begin{equation}
\vect{\alpha}_{t+1}^{\beta=0} = \begin{bmatrix}
\vect{\alpha}_t\\ \frac{1}{\sigma_n^2 + k_{t+1}} e_{t+1}
\end{bmatrix}.
\label{eq:update_alpha_subs1}
\end{equation}
This rule is very similar to the KLMS update \eqref{eq:klms_update}, and even identical when the learning rate is set to $\eta = 1/(\sigma_n^2 + k_{t+1})$. Note, furthermore, that $\beta=0$ indicates that the posterior covariance from Eq.~\eqref{eq:pcov_simple} reduces to
\begin{equation}
\pcov_t^{\beta=0} = \K_t.
\end{equation}
In other words, for $\beta=0$ we obtain a KLMS model that implies a fixed posterior covariance, equal to the prior covariance. 
The predictive variance, Eq.~\eqref{eq:beta_sigma}, simplifies to
\begin{equation}
\hat\sigma^2_{ft+1}|^{\beta=0} = k_{t+1}.
\label{eq:pred_cov_beta0}
\end{equation}
Under the adopted framework, this is the model that underlies algorithms such as KLMS from \cite{liu2008kernel} and QKLMS from \cite{chen2012quantized}.

By setting $\beta=1$ in Eq.~\eqref{eq:update_alpha_subs}, the update rule becomes
\begin{equation}
\vect{\alpha}_{t+1}^{\beta=1} = \begin{bmatrix}
\vect{\alpha}_t\\ 0
\end{bmatrix} + \frac{e_{t+1}}{\sigma_n^2 + k_{t+1} + \|\kt\|^2} \begin{bmatrix}
\kt\\ 1
\end{bmatrix}
\label{eq:update_alpha_subs2}
\end{equation}
which is very similar to the update of KNLMS, shown in Eq.~\eqref{eq:knlms_update}. The implied posterior covariance then reads
\begin{equation}
\pcov_t^{\beta=1} = \K_t \K_t + \K_t,
\end{equation}
and the predictive covariance from Eq.~\eqref{eq:beta_sigma} now becomes
\begin{equation}
\hat\sigma^2_{ft+1}|^{\beta=1} = k_{t+1} + \kt^\top \kt.
\label{eq:pred_cov_beta1}
\end{equation}
Interestingly, this KLMS model implies a predictive covariance, for each point in space, that is larger than the prior covariance $k_{t+1}$. Furthermore, a closer inspection of the second term in Eq.~\eqref{eq:pred_cov_beta1} shows that the predictive variance grows as more training data is processed. 
According to the GP framework, this is the model that underlies type-II KLMS algorithms, in particular KNLMS from \cite{richard2009online}, for which $\beta = 1$, and in general any algorithm that corresponds to $\beta>0$.
%

\begin{figure}[t!]
\centering
\includegraphics[width=\linewidth]{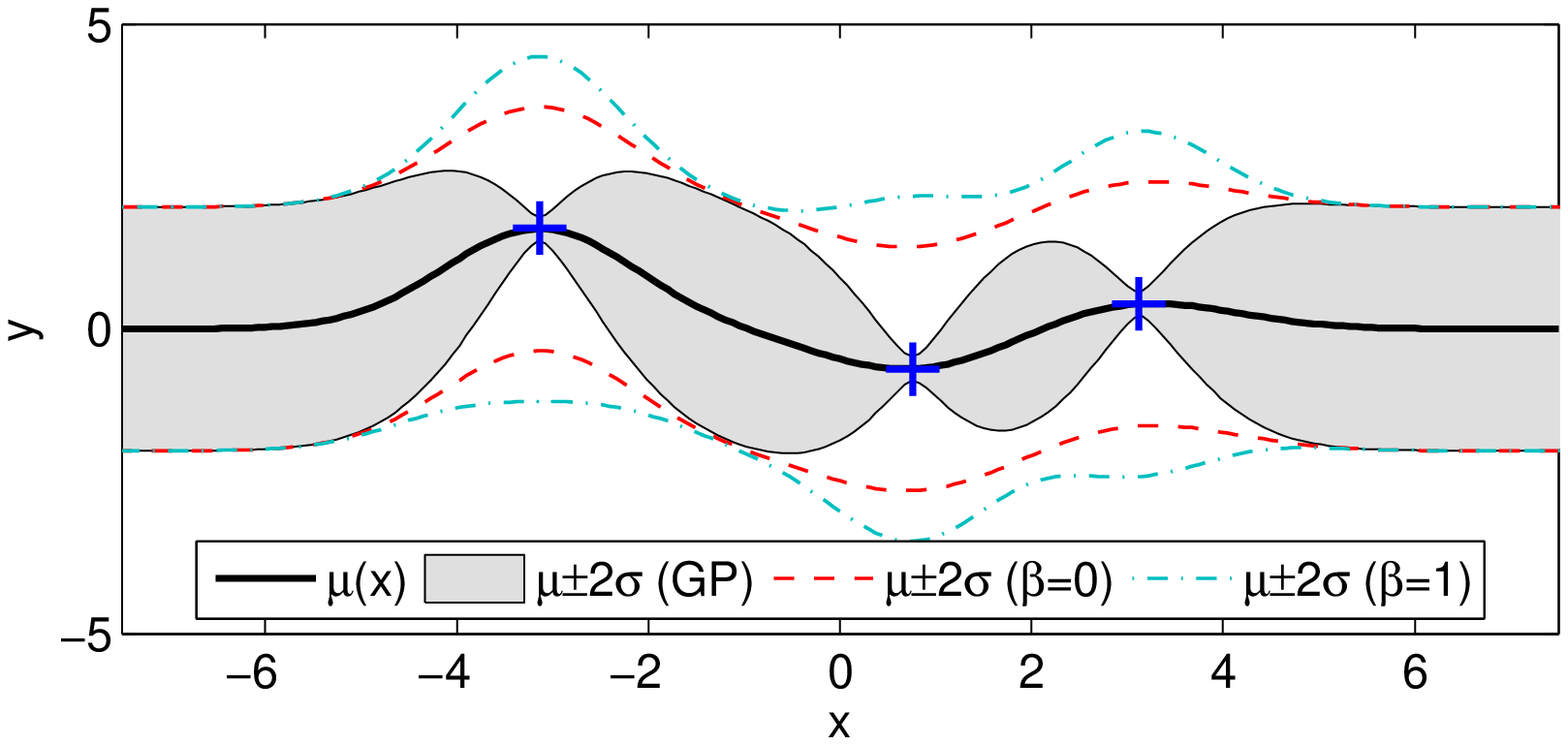}
\includegraphics[width=\linewidth]{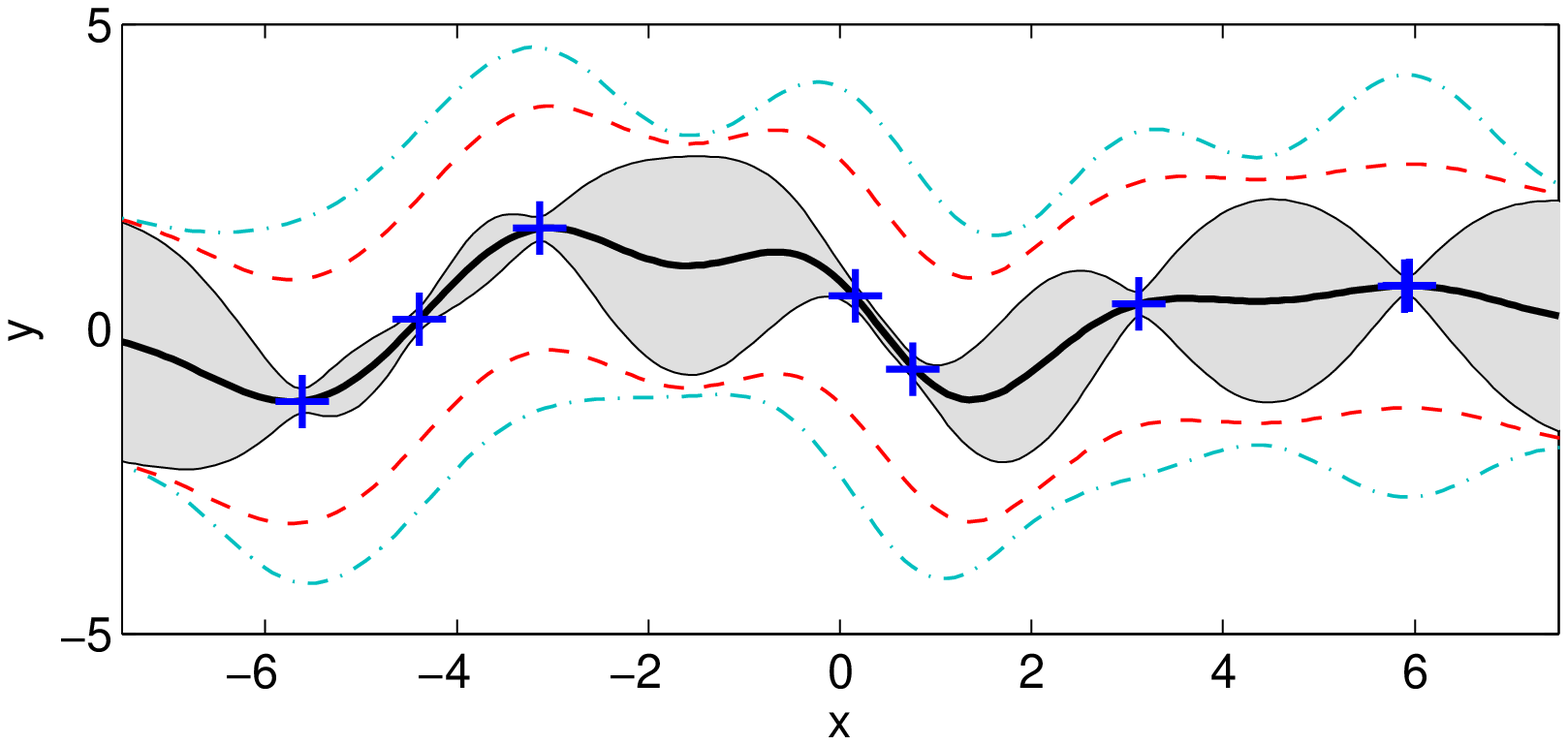}
\includegraphics[width=\linewidth]{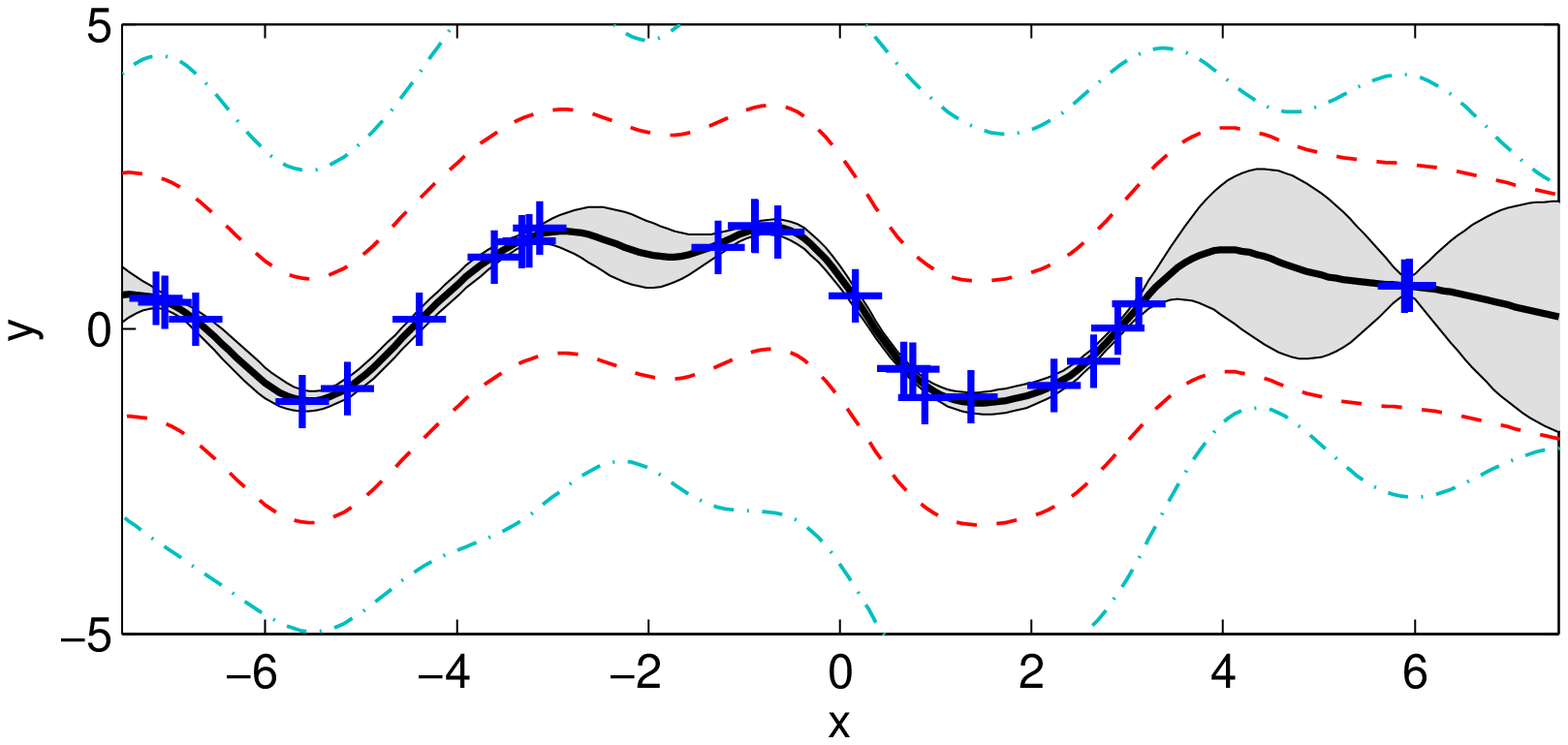}
\caption{Comparison of the predictive variance of three algorithms, for $3$ data points (top plot, data is marked as blue crosses), $8$ data points (middle), and $25$ data points (bottom). The predictive mean of GP regression is indicated as the black curve that passes through the observations. The grey zone marks the GP mean plus/minus two standard deviations $\hat \sigma_{y}$, corresponding to the GP's $95\%$ confidence interval. The dashed line marks the confidence interval for type-I KLMS ($\beta=0$), and the dash-dot line marks the confidence interval for type-II KLMS with $\beta=1$.
}
\label{fig:uncertainty}
\end{figure}

As seen through the adopted probabilistic perspective, the capability of Type-II KLMS algorithms to update all coefficients with each new data point is related to an increase in prediction uncertainty. 
These effects are illustrated in Fig. \ref{fig:uncertainty}, which compares the interpreted predictive uncertainty for the different algorithms. GP regression exhibits the expected behavior, i.e. uncertainty shrinks around the observed data points. Type-I KLMS ($\beta=0$) assumes a fixed predictive covariance, regardless of the number of observed data and the distances between them. The behavior of type-II KLMS (for instance with $\beta=1$) is rather counterintuitive: its update procedure increases the predictive variance as more data points are observed, and this happens in the neighborhoods of those points. While an in-depth study is needed to extract solid conclusions from this observation, it is interesting to note that a similar behavior has been observed in the linear recursive least squares (RLS) algorithm with forgetting factor, see \cite{vanvaerenbergh2012kernel}.

%





\section{Experiments}
\label{sec:exp}
We illustrate the relationship between the discussed algorithms through a set of numerical experiments. We used the Matlab implementations found in the KAFBOX toolbox \cite{vanvaerenbergh2013comparative}. The code for these experiments is available at \url{http://gtas.unican.es/people/steven}.

\subsection{Online regression on stationary data}

In the first experiment, we wish to study the effect of different values of $\beta$ in the KLMS algorithm from Eq.~\eqref{eq:update_alpha_subs}, which we will denote by $\beta$-KLMS. We evaluate this algorithm and three established kernel adaptive filtering algorithms on the stationary KIN40K benchmark.\footnote{Available at \url{http://www.cs.toronto.edu/~delve/data/datasets.html}} This data set is obtained from the forward kinematics of an $8$-link all-revolute robot arm, and it represents a very difficult regression problem. We randomly select $5000$ data points for online training, and $5000$ points for testing the regression.

The algorithms are considered in their evergrowing version here, in order to highlight only the influence of $\beta$. The KRLS-T algorithm, however, which implements the full online GP regression, is given a limited memory of $500$ bases, for computational reasons. A Gaussian kernel was used for all algorithms, with parameters determined offline by standard GP regression, as detailed in \cite{vanvaerenbergh2012kernel}.

\begin{figure}[t]
\centering
\includegraphics[width=\linewidth]{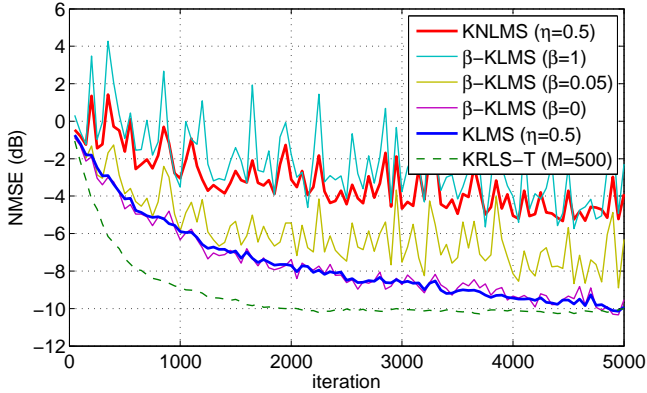}
\caption{Performance comparison for online prediction on the KIN40K benchmark regression problem.} \label{fig:kin40k}
\end{figure}

The results are shown in Fig.~\ref{fig:kin40k}. Each point of the learning curves corresponds to the test error on the entire test set. We observe that for $\beta=1$, the $\beta$-KLMS algorithm obtains similar performance to KNLMS from \cite{richard2009online}. For $\beta=0$, the performance is almost identical to that of KLMS from \cite{liu2008kernel}.

\subsection{Online prediction of time series with a model switch}

\begin{figure}[t]
\centering
\includegraphics[width=\linewidth]{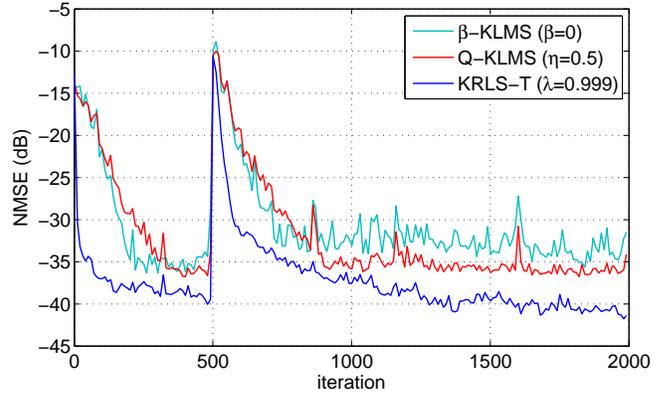}
\caption{Performance comparison for online prediction on a nonlinear channel with an abrupt change.} \label{fig:reconvergence}
\end{figure}

In Fig. \ref{fig:reconvergence} we repeat a standard experiment from the kernel adaptive filtering literature. In particular, we consider a nonlinear system comprised of a linear channel followed by a nonlinearity, and the online regression task consists in predicting the next sample of the time series produced at its output. An abrupt change in the linear channel is triggered after $500$ time steps, in order to test the algorithms' ability to reconverge. The experiment is repeated $5$ times with random channel coefficients, and the average results are shown in Fig. \ref{fig:reconvergence}. The same kernel was used for each algorithm. 

In this experiment we observe that, as is usual in the LMS and KLMS literature, the $\beta$-KLMS model exhibits tracking behavior although it is based on a static data model. Furthermore, it does so without the need of including an additional parameter to control the update step size.



%
%


\section{Conclusions}
\label{sec:concl}
We studied the connections between online GP regression and KLMS, from a probabilistic perspective. We proposed a parametric model for fixing the posterior covariance that, when plugged into the update equations for online GP regression, yields the well-known equations of several KLMS algorithms. This approach allowed us to analyze the way in which KLMS algorithms implicitly handle uncertainty. 

We furthermore categorized existing KLMS algorithms into two classes, depending on which coefficients they update during online operation: Type-I KLMS algorithms concentrate all novelty into a single coefficient, while type-II algorithms spread the novelty over many coefficients. According to the adopted probabilistic perspective, the former fixes its uncertainty regarding new data while the uncertainty of the latter grows as more data are processed. 


Finally, while the proposed parametric model shows interesting features as a KLMS algorithm, it has several aspects, such as the appropriate choice of the $\beta$ parameter, that require a further analysis.
%
Furthermore, the use of GPs as models for kernel adaptive filters could open the door to more sophisticated low-complexity algorithms, for instance by considering pseudo-inputs \cite{hensman2013gaussian}.


\bibliographystyle{IEEEbib}
\bibliography{refs}


\end{document}